\documentclass[10pt,twocolumn,letterpaper]{article}

\usepackage{iccv}
\usepackage{times}
\usepackage{epsfig}
\usepackage{graphicx}
\usepackage{amsmath}
\usepackage{amssymb}

\usepackage[pagebackref=true,breaklinks=true,letterpaper=true,colorlinks,bookmarks=false]{hyperref}

\iccvfinalcopy 

\def\iccvPaperID{****} 

\ificcvfinal\fi

\begin{document}

\title{Multi-Target Domain Adaptation via Unsupervised Domain Classification for Weather Invariant Object Detection}

\author{Ting Sun \and Jinlin Chen \and Francis Ng   \\
{\tt\small tsun@connect.ust.hk, jinlin.chen@connect.polyu.hk, francis.ng@huawei.com}
}

\maketitle
\ificcvfinal\thispagestyle{empty}\fi

\begin{abstract}
   Object detection is an essential technique for autonomous driving.  The performance of an object detector significantly degrades if the weather of the training images is different from that of test images.  Domain adaptation can be used to address the domain shift problem so as to improve the robustness of an object detector. However, most existing domain adaptation methods either handle single-target domain or require domain labels.  We propose a novel unsupervised domain classification method which can be used to generalize single-target domain adaptation methods to multi-target domains, and design a weather-invariant object detector training framework based on it.  We conduct the experiments on Cityscapes dataset and its synthetic variants, \ie foggy, rainy, and night. The experimental results show that the object detector trained by our proposed method realizes robust object detection under different weather conditions.
\end{abstract}

\section{Introduction}
\label{sec:introduction}
Object detection is a fundamental computer vision task which is widely used in many real-life applications. A typical image object detector takes in an image and outputs the class labels and location bounding box coordinates of all the objects of certain categories in the image.  This technique plays an essential role in autonomous driving.  The detection results can be used to intelligentize safety driving and facilitate auto navigation \etc.

Just like in other computer vision tasks, deep learning \cite{LeCun2015,Schmidhuber2015,movable2019} approaches have achieved excellent performance on the object detection benchmark datasets \cite{Liu_2020_CVPR,9308604}. However, in the real application of autonomous driving, the variant weather conditions causes a considerable domain shift between the training and test images, and consequently degrades the performance of a well-trained object detector \cite{Gopalan2011}.  A straightforward solution is to collect more training data that cover all possible weather conditions.  Although it is not difficult to collect a large number of raw images, manual annotations are laborious and expensive to obtain.

Our problem setting is that during training, images following source domain distribution are provided with full annotations, while images following multiple target domain distributions are available without any annotations.  To avoid the cost of annotating each new target domain, domain adaptation (DA) approaches aim to transfer knowledge from source domain(s) to unlabeled target domain(s) to reduce the discrepancy between their distributions, typically by exploring domain-invariant data structures.  Most existing domain adaptation methods focus on single target domain, yet in autonomous driving, there are several typical weather conditions, \eg foggy, rainy and night \etc.

In this paper, without using domain labels, we propose a multi-target-domain adaptation method based on unsupervised domain classification for weather invariant object detection.  Our method first trains an style transfer model \cite{Huang_2018_ECCV} between the source domain and mixed target domain, which contains all different weather conditions.  This style transfer model is used to extract style features from all the target domain images.  We use k-means \cite{1056489,ilprints778} to conduct unsupervised clustering so that the target domain images are classified based on their weather conditions.  Then for each clustered distinct target domain, a new source-target domain style transfer model is trained.  These style transfer models are used to generate annotated target domain images from the source domain images.  Thus an augmented training dataset which contains all weather conditions are generated.  A weather-invariant object detector is trained on this augmented dataset in a supervised manner.

Our contribution in this work is twofold:
\begin{itemize}
	\item we propose a novel unsupervised domain classification method which can be used to generalize single-target domain adaptation methods to multi-target domains. 
	\item we successfully design and test a weather-invariant object detector training framework via our proposed unsupervised multi-target domain adaptation method.  
\end{itemize}

The reminder of this paper is organized as follows.  Sec.~\ref{sec:related_work} reviews some previous work on object detection, style transfer and domain adaptation.  Preliminaries are briefed in Sec.~\ref{sec:preliminaries}. Our proposed method is presented in Sec.~\ref{sec:proposed_method} which is followed by experimental results in Sec.~\ref{sec:experiments}.  Sec.~\ref{sec:conclusion} concludes the paper.
 
\begin{figure*}[h]
	\centering
	\includegraphics[width=\textwidth]{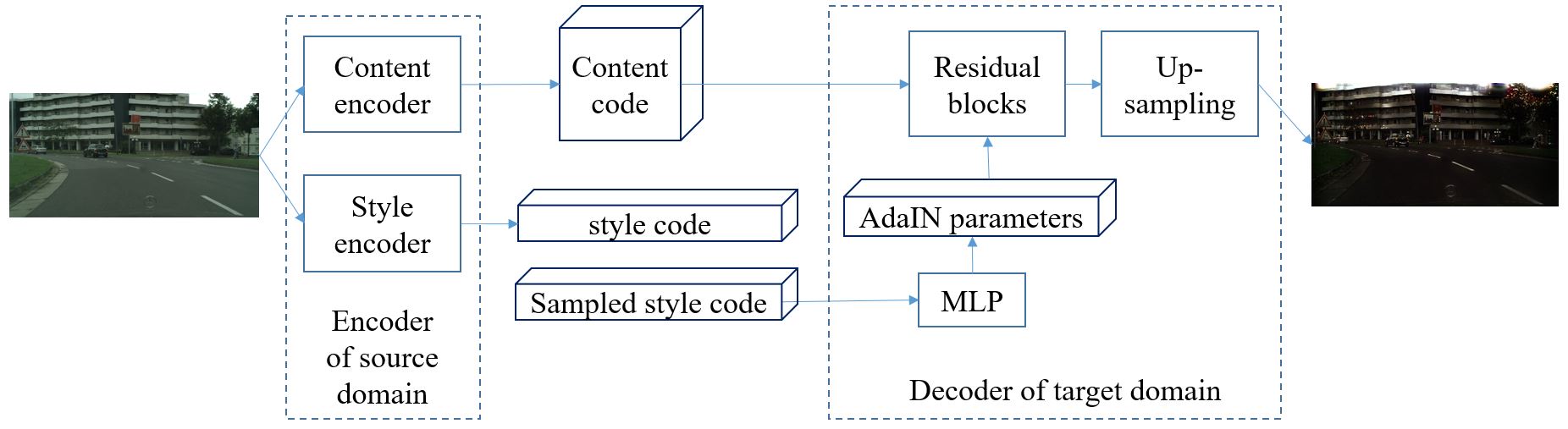}
	\caption{The auto-encoder architecture of MUNIT \cite{Huang_2018_ECCV}.  This figure shows an example of the style transfer process, \ie transfering a daytime street view image into a night one.  The content of the daytime image is extracted by the encoder of the daytime domain, and the night image is generated by the decoder of the night domain with the same content and a randomly sampled style code.}
	\label{fig:MUNIT_transfer}
\end{figure*}

\section{Related work}
\label{sec:related_work}

\subsection{Object detection}
Object detection is a longstanding and fundamental computer vision problem \cite{liuli2020}, and it has a wide range of application such as autonomous driving.  A typical object detector is to determine whether there are any objects of given categories existing in the input image, and output their location bounding boxes and category labels.  The traditional work usually handle object detection by sliding window approaches \cite{Dalal2005,liuli2020}.  Since deep learning \cite{LeCun2015,Schmidhuber2015}, especially convolutional neural networks (CNN or ConvNet) \cite{LeCun1989} has emerged as a powerful tool that focuses on learning features or representations directly from raw images in a hierarchical manner \cite{sun2017}, and has become the most prevalent object detection method that achieves state-of-the-art performance \cite{Liu_2020_CVPR,9308604}.  Existing deep-learning-based object detection frameworks can be grouped into two types \cite{liuli2020}, depending on whether a preprocessing step for generating object proposals is required (region based) or not.  The region-based CNNs have dominated object detection since R-CNN \cite{Girshick2014}, and the leading results on popular benchmark datasets are mostly based on Faster R-CNN \cite{Ren2016}, which is also flexible to modify for other purposes.  Our domain adaptation method is tested on Faster R-CNN \cite{Ren2016}, and we generalize its ability in multiple target domains. 

\subsection{Style transfer}
Style transfer aims at modifying the style of an image while preserving its content \cite{Huang_2018_ECCV}.  In our case different styles of an image means the same street-view under different weather conditions.  Early style transfer models require paired images for training \cite{Isola_2017_CVPR}, or can only conduct deterministic \cite{Zhu_2017_ICCV,pmlr-v70-kim17a,Taigman2016} or unimodal mapping \cite{Liu2017}.  A multimodal unsupervised image-to-image translation (MUNIT) framework was proposed in \cite{Huang_2018_ECCV}.  However, these methods are designed for single source domain and single target domain pair, but in autonomous driving, there are several typical different weather conditions, \ie foggy, rainy and night \etc.  StarGAN \cite{Choi_2018_CVPR} is an approach for multi-domain image-to-image translation, but it requires domain labels for training.  Our domain adaptation method adopts MUNIT \cite{Huang_2018_ECCV} for style feature extraction and data augmentation.

\subsection{Domain adaptation for object detection}
Supervised machine learning methods assume that training and test data are sampled \textit{i.i.d.} from the same distribution, but in practice their distributions may differ, \ie domain shift exists \cite{Quionero2009,Peng2019}.  To avoid the cost of annotating each new test domain, domain adaptation (DA) approaches aim to reduce the discrepancy between the distributions of training and test data, typically by exploring domain-invariant data structures.  

There are two main groups of domain adaptation methods.  The first group try to align the extracted features by designing losses or using generative adversarial networks(GANs) \cite{Goodfellow2014} to encourage the confusion between source and target domains \cite{9308604}.  The second group makes use of style transfer model to augment the training dataset \cite{liuli2020}.  Our proposed method falls into the second group.  DA has been widely studied for image classification \cite{pmlr-v97-peng19b}, and the first end-to-end trainable object detection model is proposed in \cite{Chen2018}, where the features of source domain and target domain are aligned from image-level and instance-level by adversarial training. X. Zhu \etal  \cite{Zhu2019} proposed to mine the discriminative regions and focus on aligning them.   The idea of strong local alignment and weak global alignment is contributed in \cite{Saito2019}, which focuses the adversarial alignment loss on images that are globally similar and puts less emphasis on aligning images that are globally dissimilar.  A hierarchical domain feature alignment model is proposed in \cite{He2019}, with an information invariant scale reduction module for promoting training efficiency and a weighted gradient reversal layer for characterizing hard confused domain samples.  The instance alignment in \cite{Zhuang2020} is category-aware.  As mentioned previously, image transfer models are commonly used to generate annotated target domain images, so that the object detector can be trained in a supervised manner in the generated target domain images \cite{8852008,Devaguptapu2019}.  To alleviate the imperfection of style translation model, feature-level alignment is applied together with the pixel-level adaptation in \cite{Shan2019,Hsu2020}.  A more complete survey can be found in \cite{9308604}. 

Our method falls into the second group, \ie using a style transfer model to augment the training dataset.  Most existing domain adaptation methods focus on single target domain, or require domain labels for straightforward extention, yet in autonomous driving, there are several typical weather conditions, \eg foggy, rainy and night \etc.  Our method achieves weather invariant object detection without using domain labels.

\section{Preliminaries}
\label{sec:preliminaries}

\subsection{Faster R-CNN}
Faster R-CNN \cite{Ren2016} is a region-based object detector that mainly consists of three components: a fully convolutional neural network as feature extractor, a region proposal network (RPN) which proposes regions of interest (ROI), and a ROI based classifier.  An input image is first represented as a convolutional feature map produced by the feature extractor, then RPN produces the probability of a set of predefined anchor boxes for containing an object or not, together with refined ROIs.  Whereafter the ROI-wise classifier predict the category labels as well as the refinements for the proposed ROIs based on the feature obtained using ROI-pooling layer.  The whole network is trained by minimizing the sum of the losses of RPN and ROI classifier:

\begin{equation}
	\mathcal{L} = \mathcal{L}_{RPN} + \mathcal{L}_{ROI}
\end{equation} 

Both $\mathcal{L}_{RPN}$ and $\mathcal{L}_{ROI}$ consists of a cross-entropy loss to penalize mis-classification and a regression loss on the box coordinates for better localization.  

\subsection{MUNIT}
A multimodal unsupervised image-to-image translation (MUNIT) framework is proposed in \cite{Huang_2018_ECCV} to generate diverse outputs from one source domain image.  A well trained MUNIT model between two domains consists of two auto-encoders, \ie a pair of encoder and decoder for each domain.  The encoder is used to decomposes an image of that domain into a content code which is domain-invariant, and a style code which captures domain-specific properties.  The decoder takes in a content code and a style code to generate an image of its style.  To translate an image from a source domain to the target domain, first use the encoder of the source domain to extract its content code, then generate an image with this content code and a randomly sampled style code using the decoder of the target domain.  An example of the style transfer process is shown in Figure~\ref{fig:MUNIT_transfer}.

It is worth mentioning that the `style code' represents the variation of a fixed-content image of a certain domain, \ie a street scene could have many possible appearance at night due to timing, lighting \etc, while the style type or domain category is determined by which decoder is used to generate the image.

\section{Proposed method}
\label{sec:proposed_method}
We consider the problem setting of multi-target domain adaptation for object detection, and denote the source domain as $\mathcal{S}$, and the mixed target domain $\mathcal{T}_{mix} = \{\mathcal{T}_1, \mathcal{T}_2, \cdots, \mathcal{T}_N \}$, which is a mixture of $N$ distinct domains.  During training we have access to the source domain images $x^{S}_{i}$ and their full annotations $y^{S}_{i}$, \ie object bounding boxes and category labels, while for target domains only raw images $x^{T}_{j}$ are provided with no bounding box, category label or domain label.  Test images can come from either $\mathcal{S}$ or one of $\mathcal{T}_{mix}$. 

\begin{figure*}[h]
	\centering
	\includegraphics[width=\textwidth]{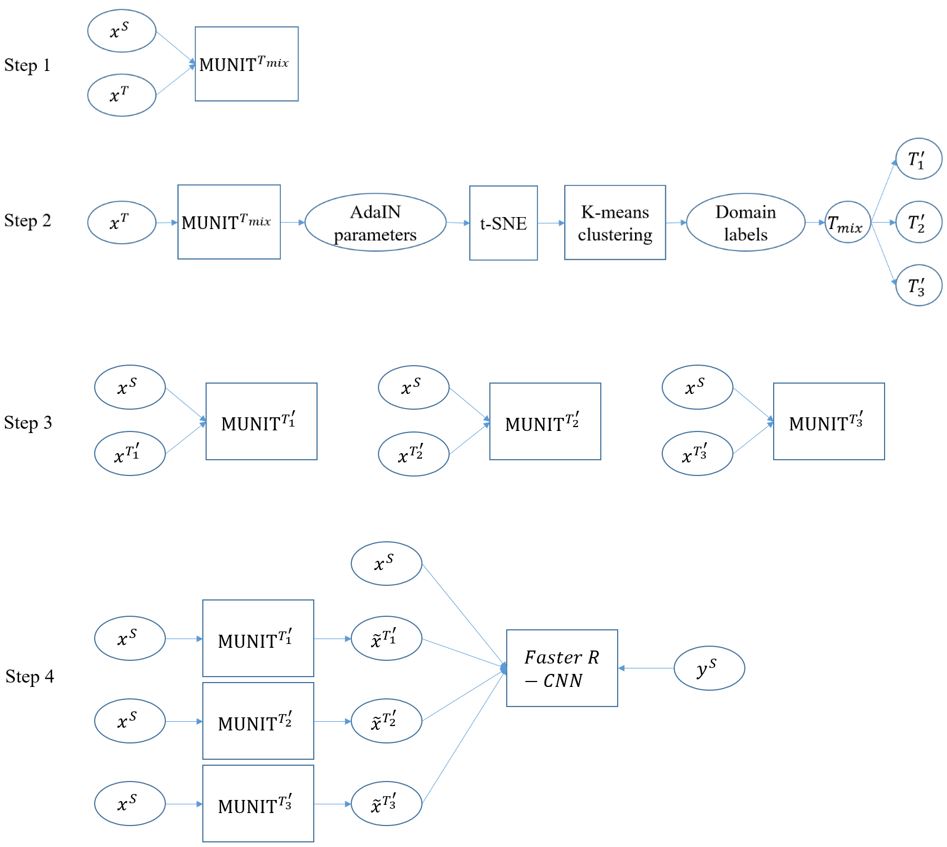}
	\caption{The four steps of our proposed training process.  Detailed description
		can be found in the text.}
	\label{fig:4_steps}
\end{figure*}

Our approach falls into the group that uses a style transfer model to augment the training images.  In order to handle multiple target domains without using domain labels, we propose an novel unsupervised domain classification method based on the style feature extracted from a style transfer model.  We adopt Faster R-CNN \cite{Ren2016} as our object detection model and MUNIT \cite{Huang_2018_ECCV} for style feature extraction and data augmentation.  The proposed method does not alter the inference process, and the training process can be divided into four steps as shown in Figure~\ref{fig:4_steps}.

In the first step a MUNIT \cite{Huang_2018_ECCV} model ($\text{MUNIT}^{\mathcal{T}_{mix}}$) is trained between $\mathcal{S}$ and $\mathcal{T}_{mix}$.  As will be seen in Sec.~\ref{sec:experiments}, this $\text{MUNIT}^{\mathcal{T}_{mix}}$ model trained on mixed target domains, without distinguishing them, cannot generate images that reflects the distribution of $\mathcal{T}_{mix}$, but it can be used to disentangle the content and style features of a target domain image $x^{T} \in \mathcal{T}_{mix}$.  

Our key contribution lies in the second step, where we use $\text{MUNIT}^{\mathcal{T}_{mix}}$ to extract the AdaIN \cite{Huang_2017_ICCV} parameters of all the target domain images $x^{T} \in \mathcal{T}_{mix}$ as their style features, and after t-SNE \cite{t-sne} dimension reduction, we conduct k-means \cite{1056489,ilprints778} to cluster the mixed target domain images into $k$ groups.  The $k$ is found based on the mean Silhouette Coefficient \cite{ROUSSEEUW198753}.  Figure~\ref{fig:4_steps} shows the case of $k = 3$.

Once the mixed target domain images are divided into $k$ groups, \ie $k$ distinct domains $\{\mathcal{T}^{'}_{1}, \mathcal{T}^{'}_{2}, \cdots \mathcal{T}^{'}_{k}\}$, based on their styles, we train another $k$ MUNIT models, one $\text{MUNIT}^{\mathcal{T}^{'}_{j}}$ between source domain $\mathcal{S}$  and separated target domain $\mathcal{T}^{'}_{j}$ pair as shown in step 3 in Figure~\ref{fig:4_steps}.

In the last training step, we use the $k$ MUNIT models from step 3 to transform the annotated source domain images into distinct annotated target domains, thus obtaining the augmented training dataset.  The object detector trained on this augmented dataset has robust performance on all source and target domains.

\begin{figure*}[!htpb]
	\centering
	\includegraphics[width=0.24\textwidth]{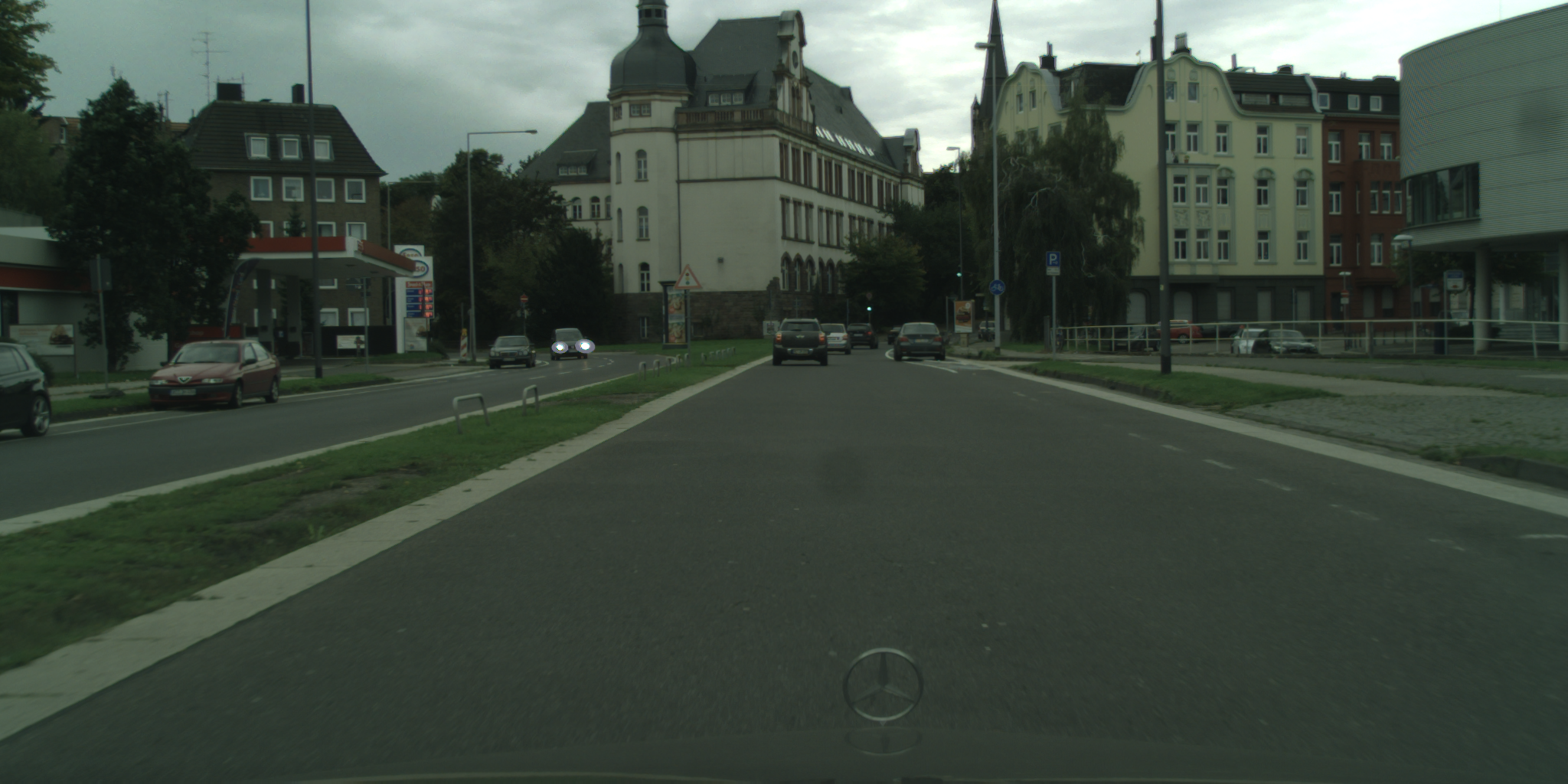}
	\includegraphics[width=0.24\textwidth]{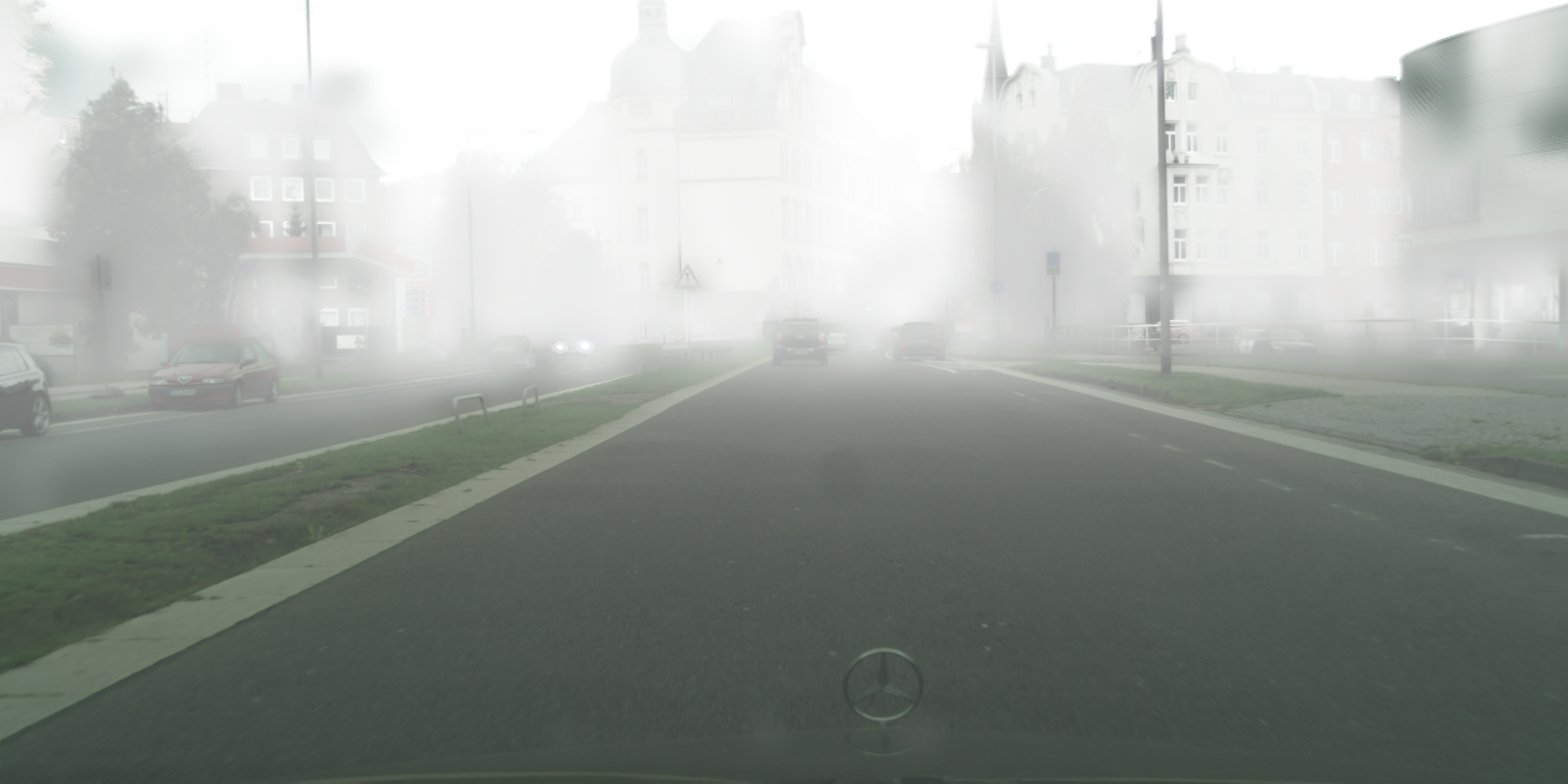}
	\includegraphics[width=0.24\textwidth]{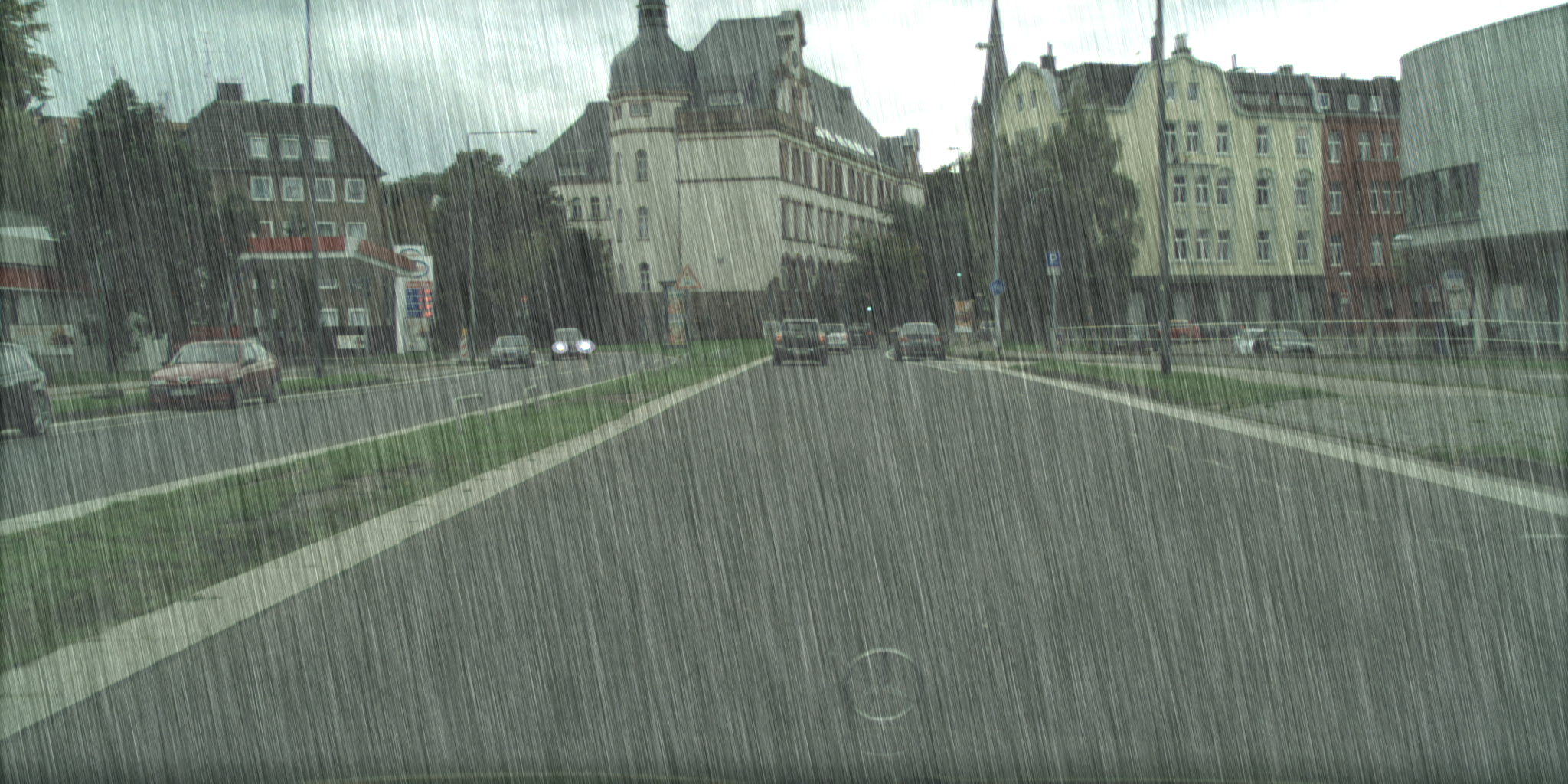}
	\includegraphics[width=0.24\textwidth]{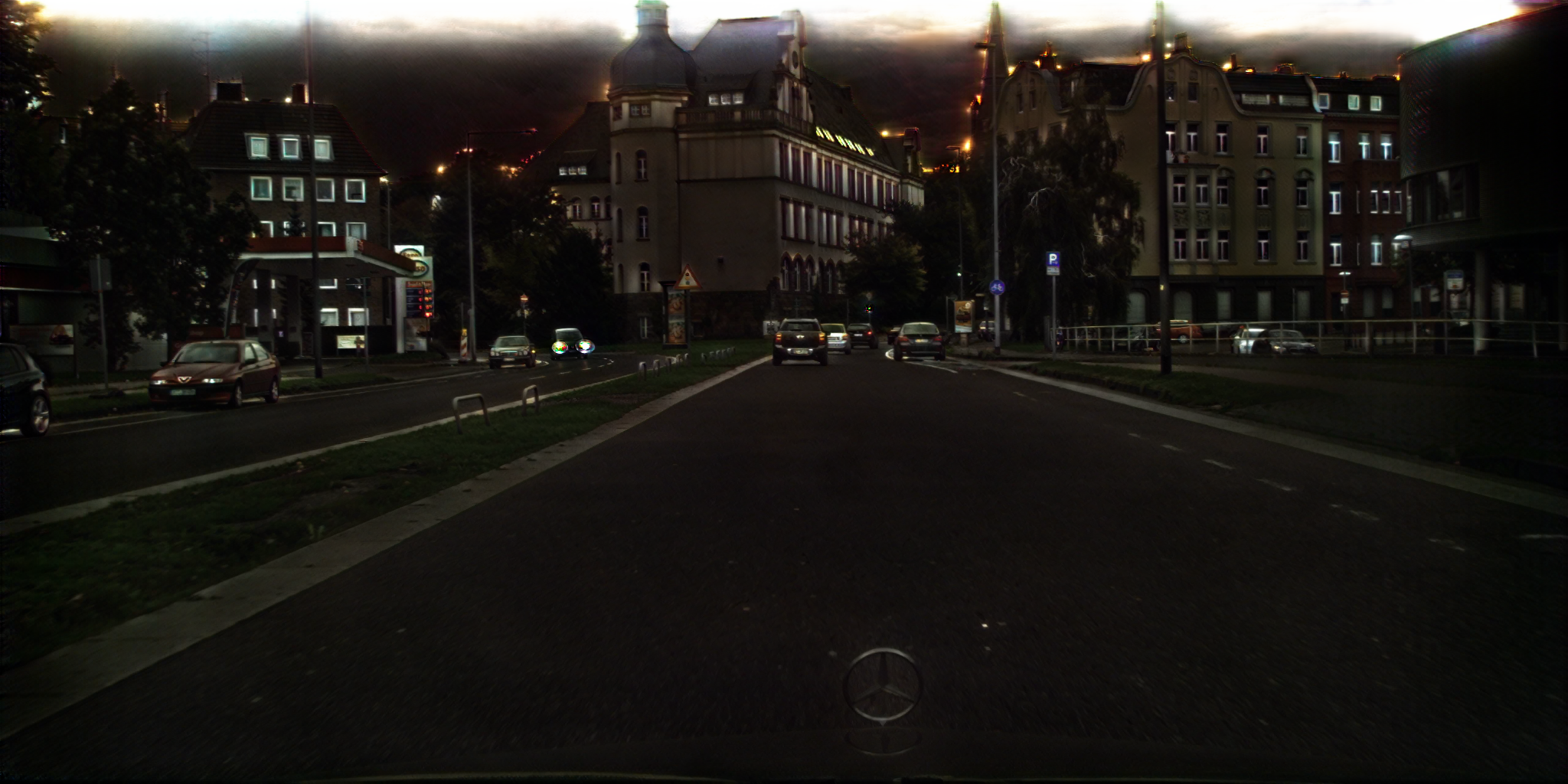} \\
	\includegraphics[width=0.24\textwidth]{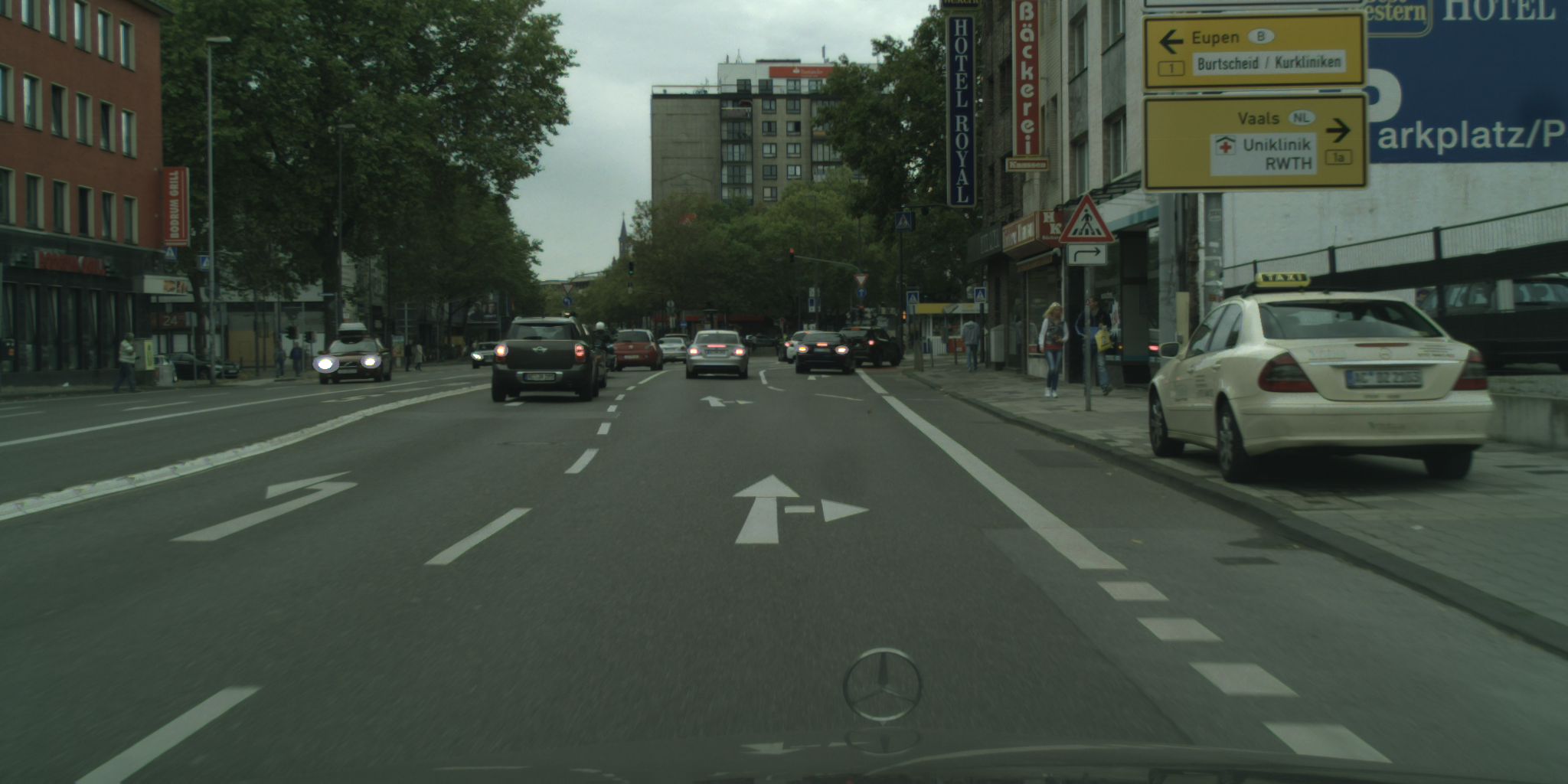}
	\includegraphics[width=0.24\textwidth]{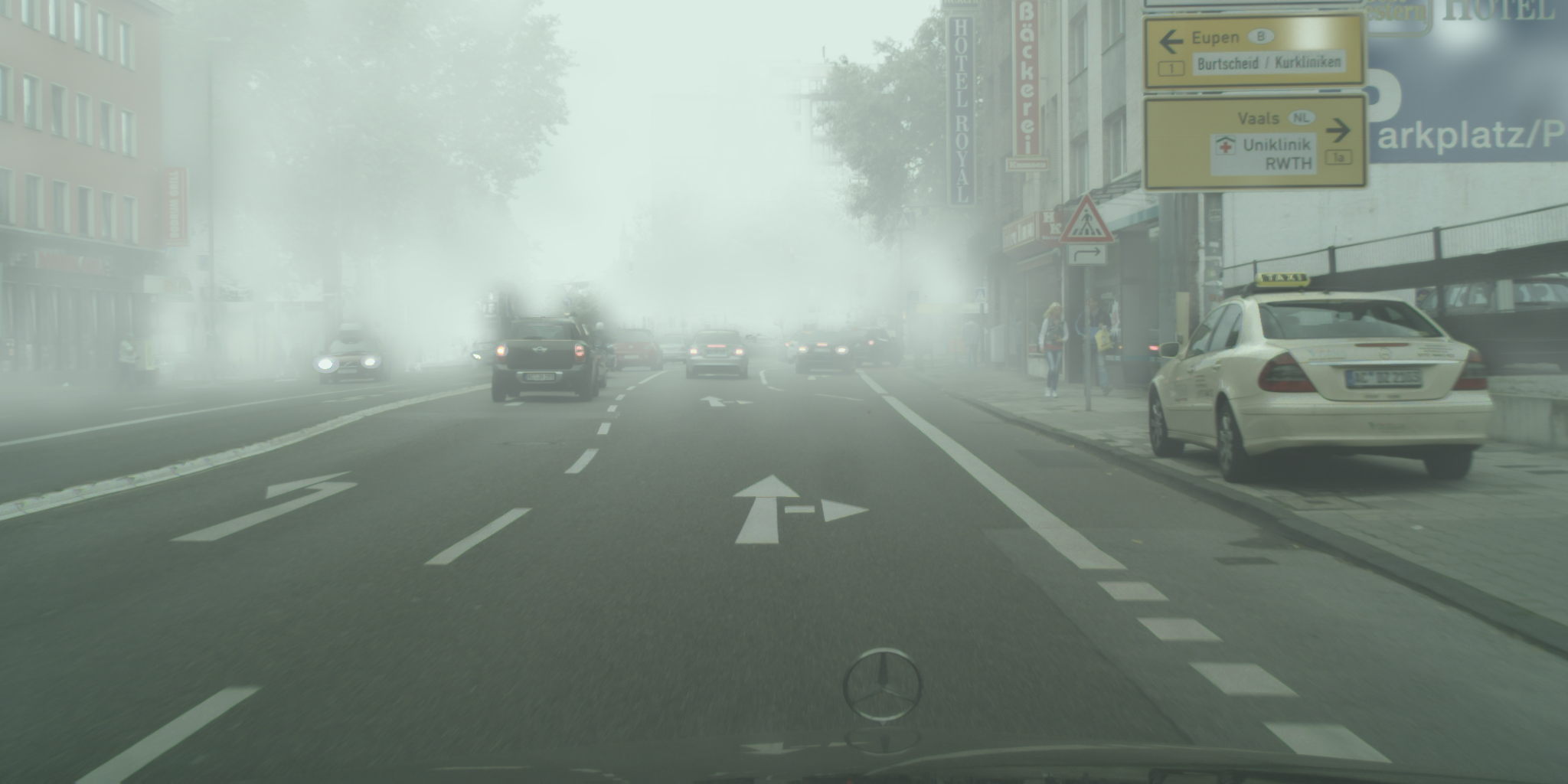}
	\includegraphics[width=0.24\textwidth]{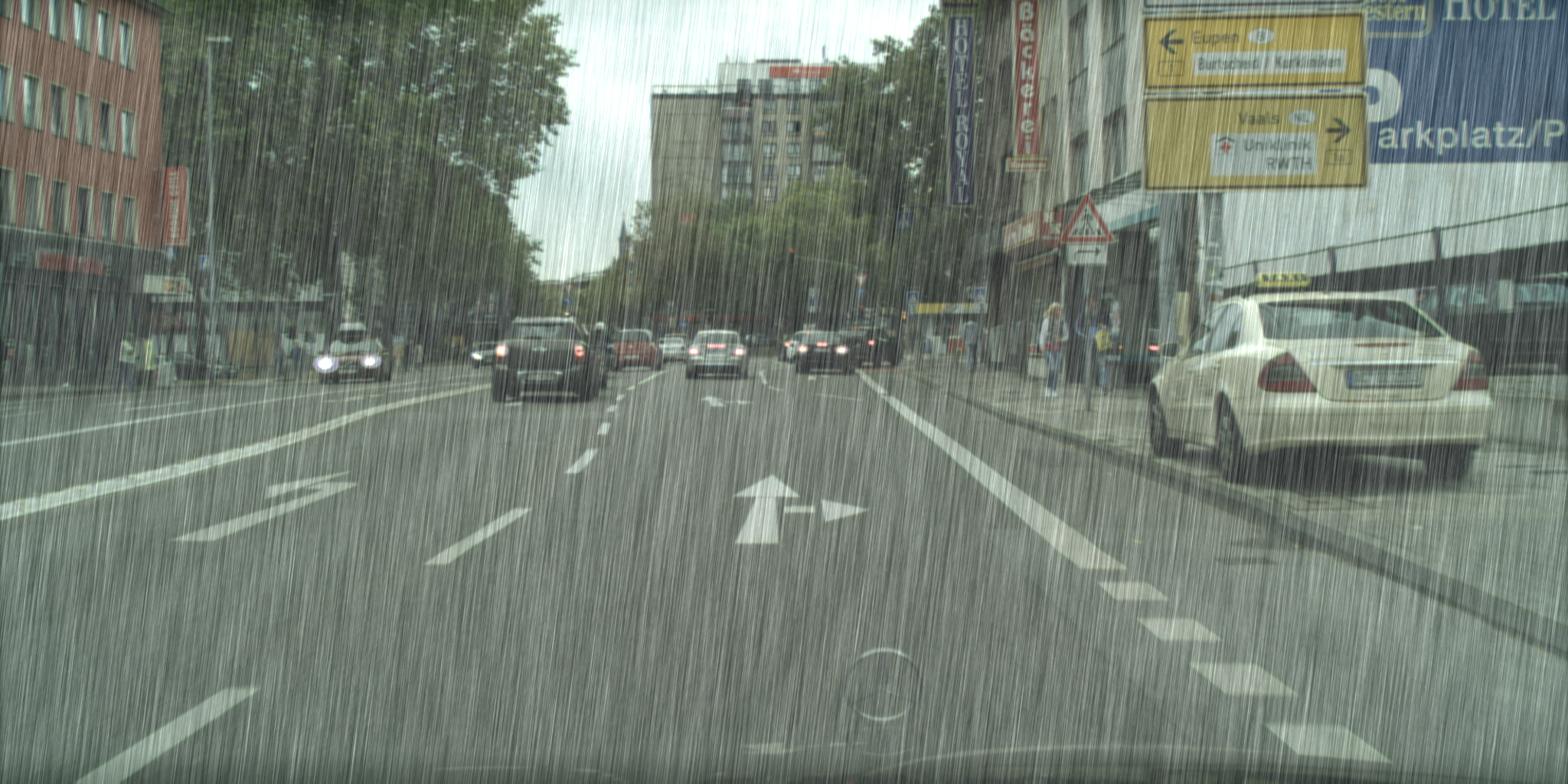}
	\includegraphics[width=0.24\textwidth]{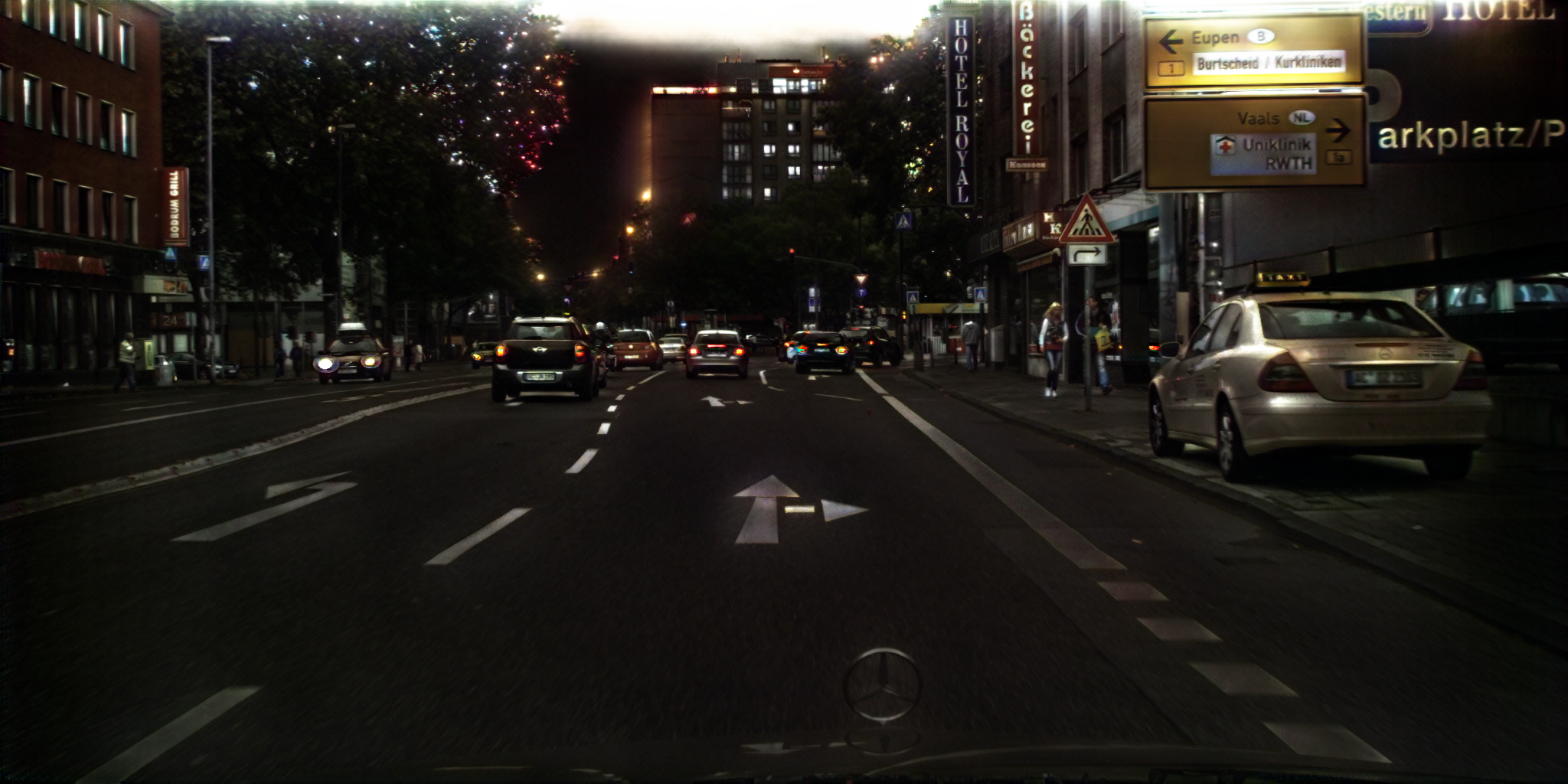} \\
	\caption{Sample images of Cityscapes \cite{Cordts2016Cityscapes} and synthesized foggy \cite{SDHV18}, rainy \cite{Li_2019_CVPR} and night \cite{night} images.}
	\label{fig:samples_images}
\end{figure*}

\begin{figure*}[!htpb]
	\centering
	\includegraphics[width=0.24\textwidth]{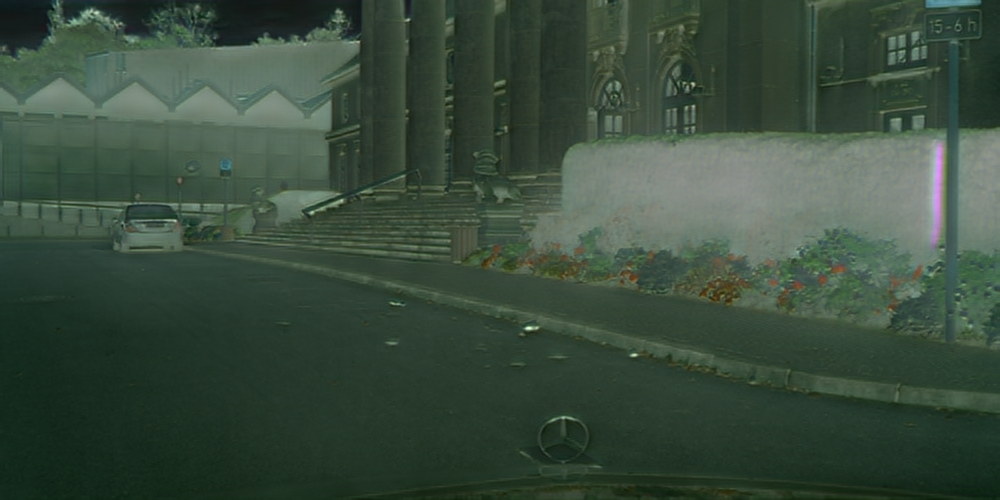}
	\includegraphics[width=0.24\textwidth]{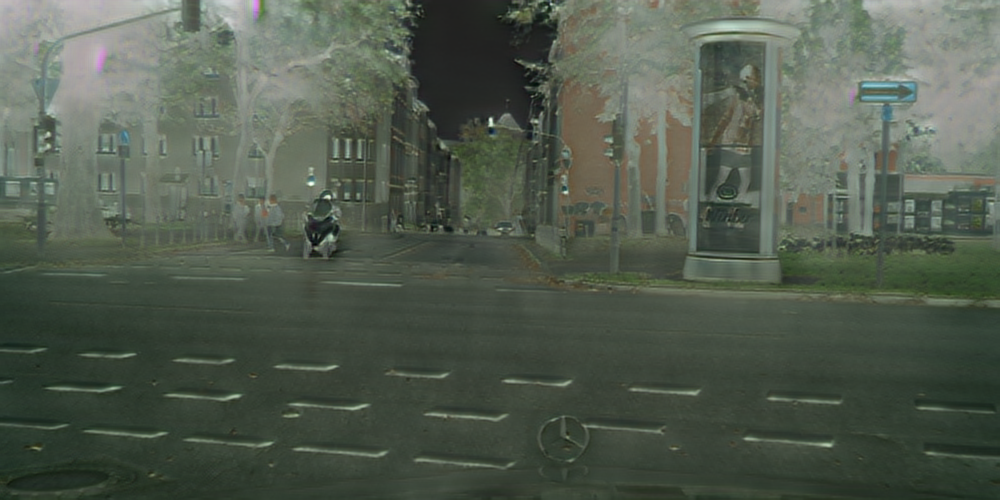}
	\includegraphics[width=0.24\textwidth]{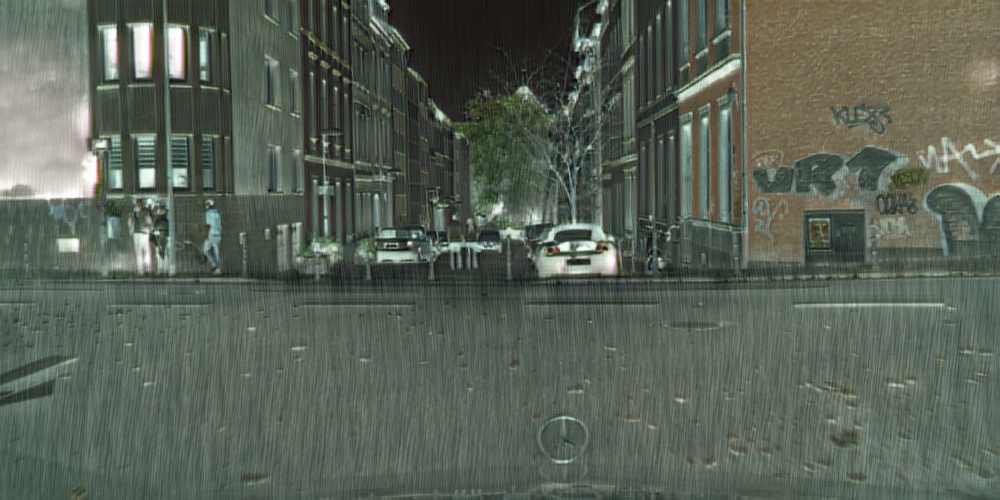}
	\includegraphics[width=0.24\textwidth]{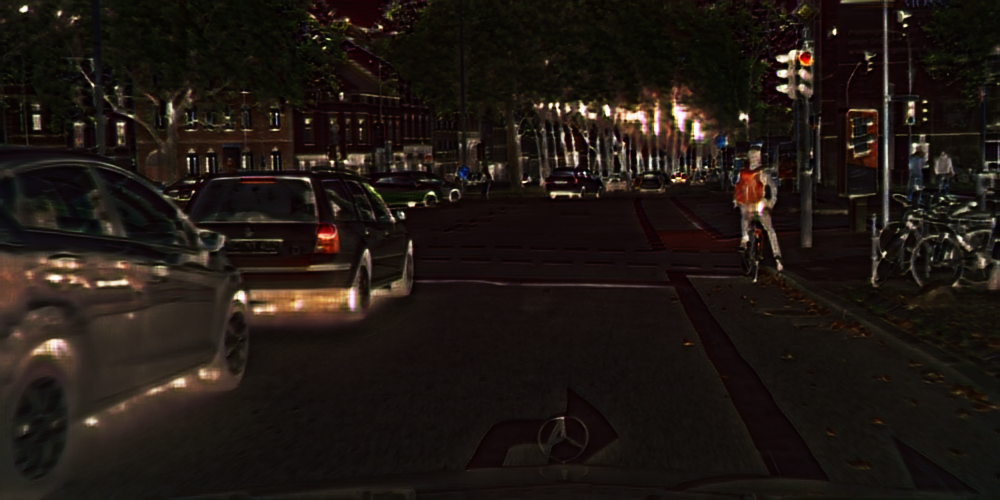} \\
	\includegraphics[width=0.24\textwidth]{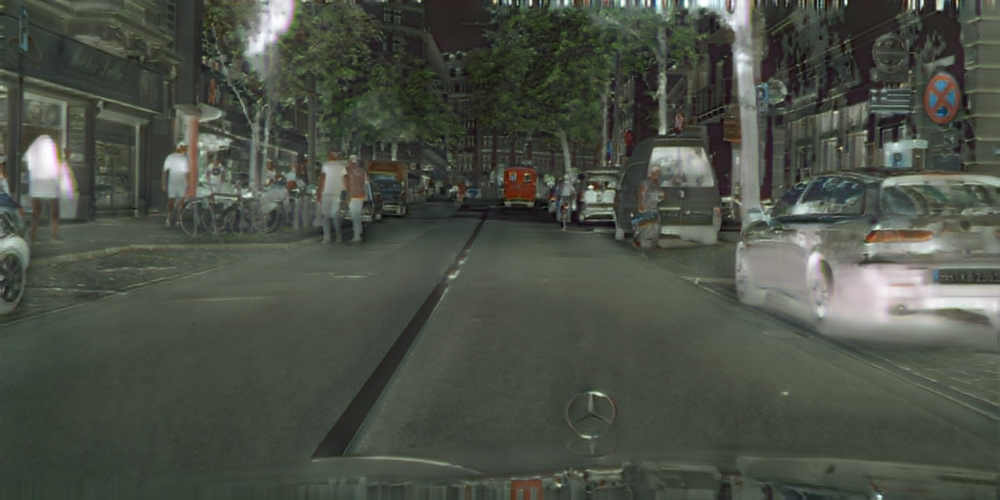}
	\includegraphics[width=0.24\textwidth]{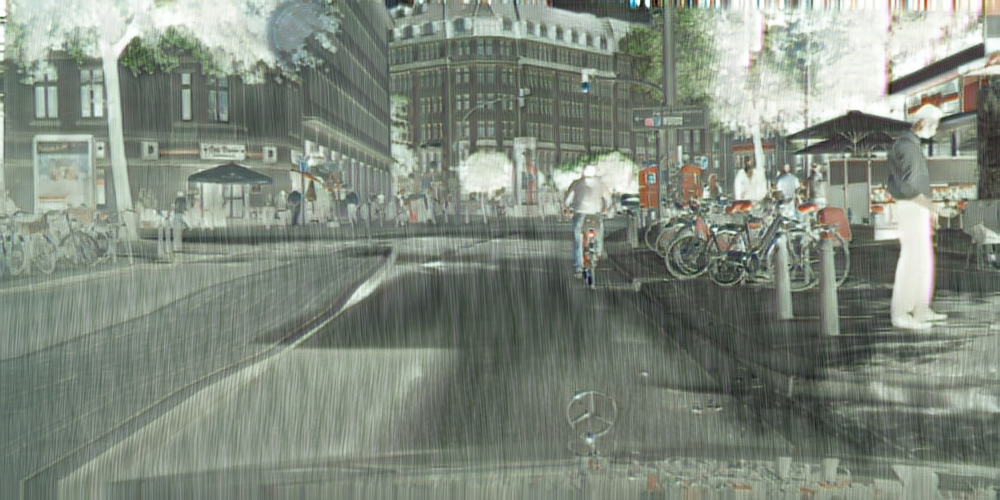}
	\includegraphics[width=0.24\textwidth]{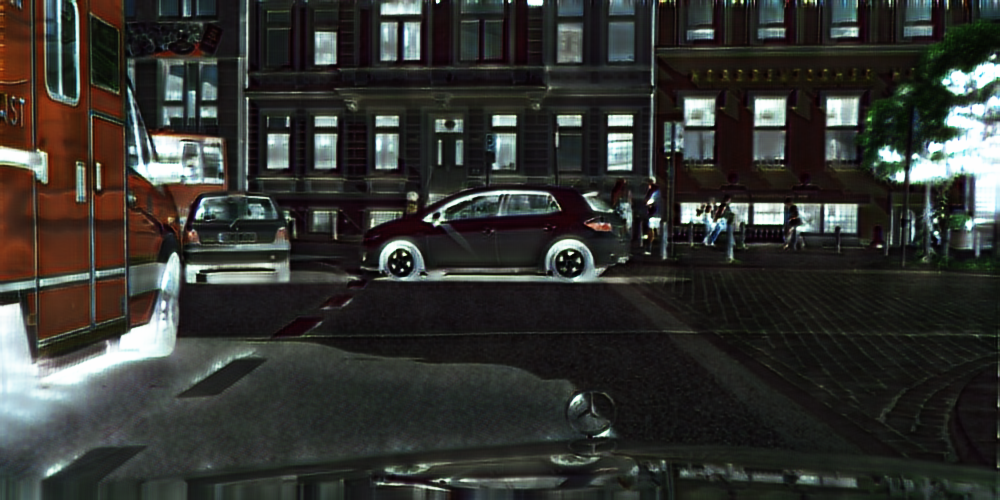}
	\includegraphics[width=0.24\textwidth]{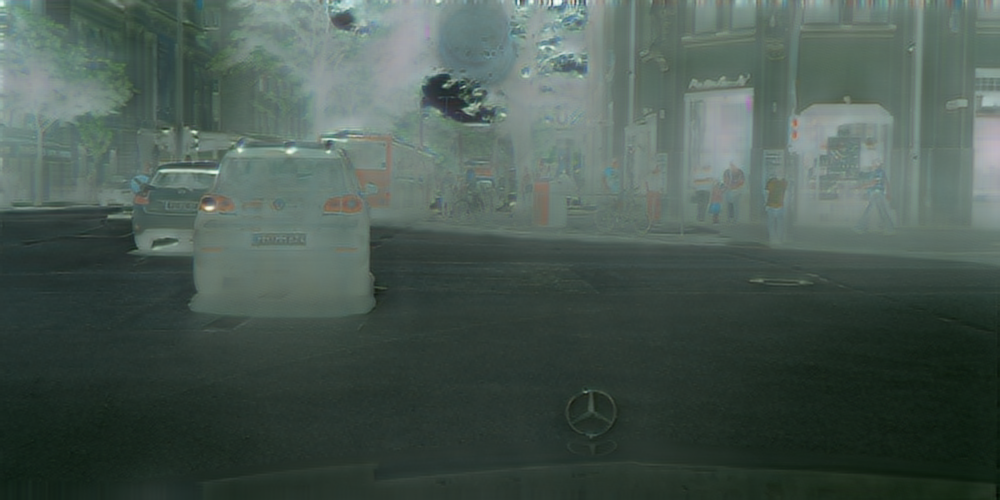} \\
	\caption{Sample target domain images generated by $\text{MUNIT}^{\mathcal{T}_{mix}}$.}
	\label{fig:naive_mix_gen_img}
\end{figure*}

\begin{figure}[!htpb]
	\centering
	\includegraphics[width=0.5\textwidth]{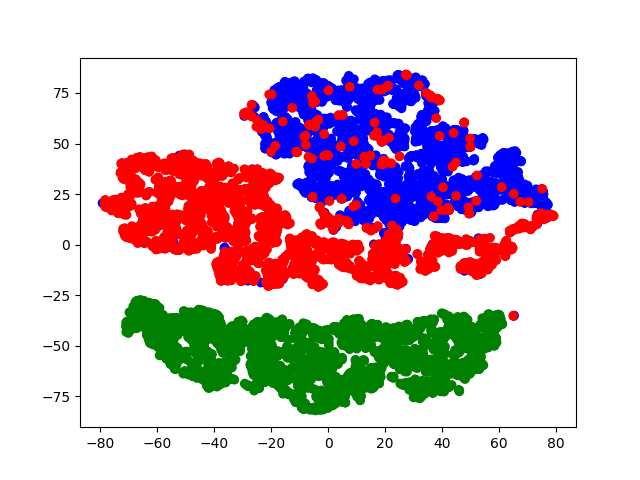}
	\caption{2D t-SNE \cite{t-sne} embedded AdaIN \cite{Huang_2017_ICCV} parameters of $\mathcal{T}_{mix}$ extracted by $\text{MUNIT}^{\mathcal{T}_{mix}}$.  The samples of rainy, night and foggy images are colored blue, green and red.}
	\label{fig:2d_adaIN_params}
\end{figure}

\begin{table}[!htpb]
	\begin{center}
		\begin{tabular}{ c | c  c c c }
			& total        & rainy (\%) & night (\%) & foggy (\%) \\
			\hline
			cluster 1 & 3,600        & 81.22     & 0.00        & 18.78      \\ 
			cluster 2 & 2,679        & 0.41      & 89.40       & 10.19      \\  
			cluster 3 & 2,646        & 1.51      & 21.92       & 76.57      \\    
		\end{tabular}
	\end{center}
	\caption{Numerical k-means \cite{1056489,ilprints778} clustering results for $k = 3$.  Each row lists the ingredient of one cluster.}
	\label{tab:clustering}
\end{table}

\begin{figure*}[!htpb]
	\centering
	\includegraphics[width=0.24\textwidth]{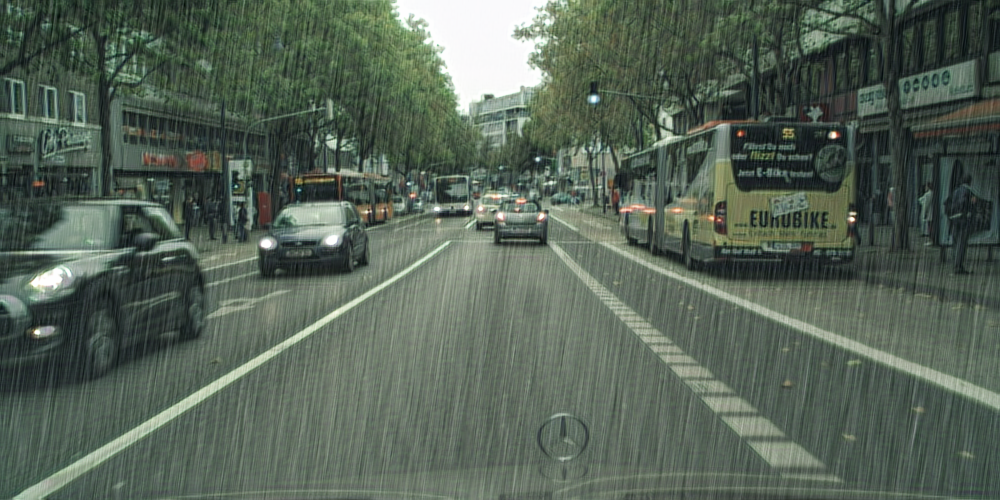}
	\includegraphics[width=0.24\textwidth]{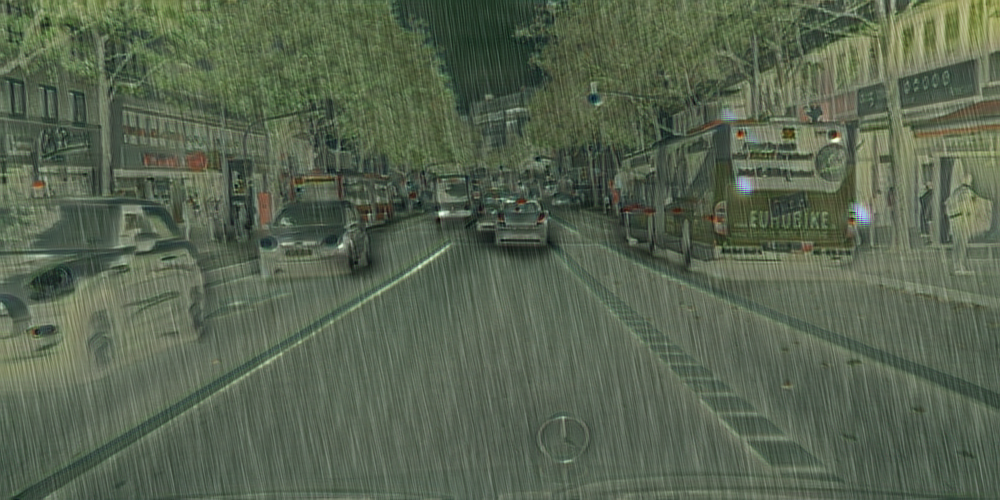}
	\includegraphics[width=0.02\textwidth]{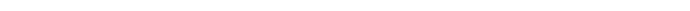}
	\includegraphics[width=0.24\textwidth]{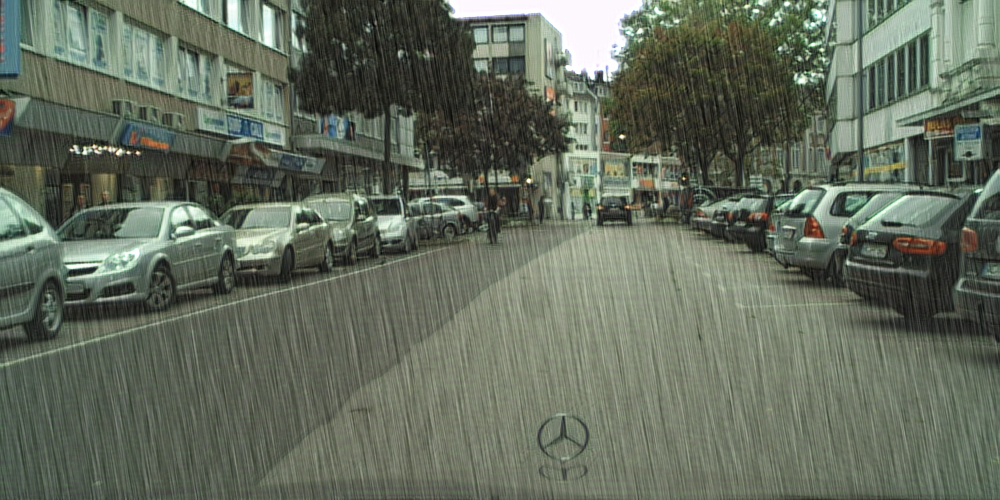}
	\includegraphics[width=0.24\textwidth]{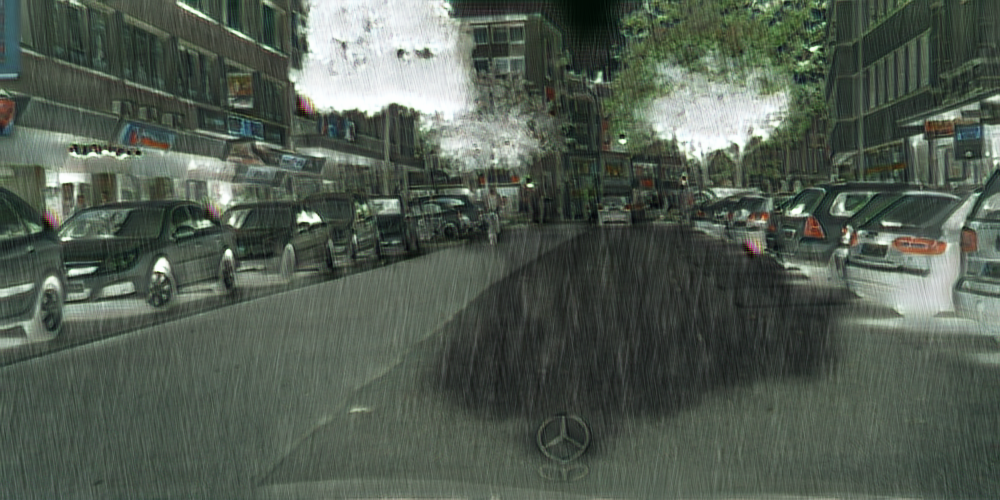} \\
	\includegraphics[width=0.24\textwidth]{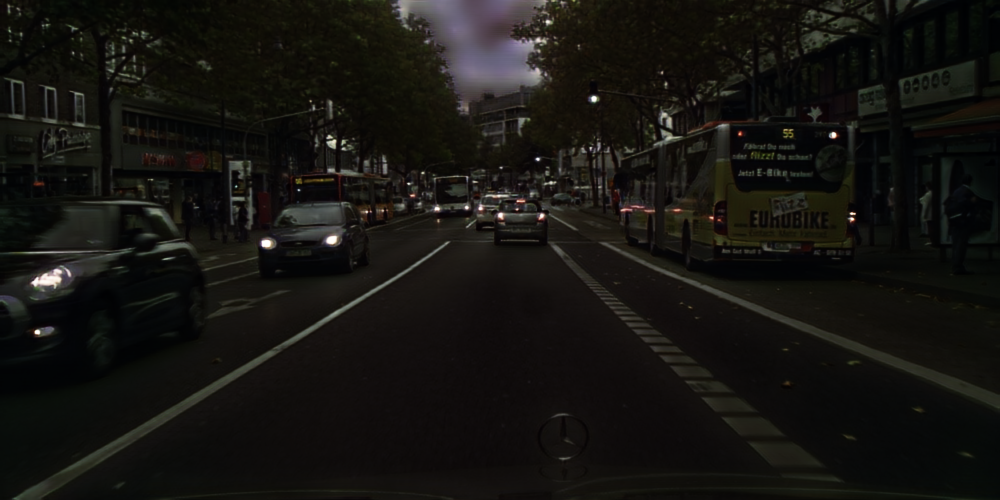}
	\includegraphics[width=0.24\textwidth]{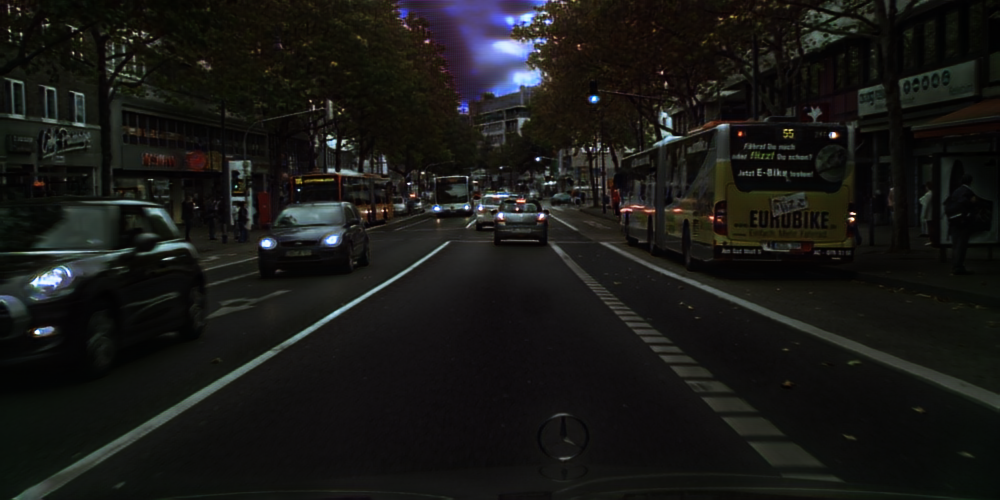}
	\includegraphics[width=0.02\textwidth]{img/white_bar.jpg}
	\includegraphics[width=0.24\textwidth]{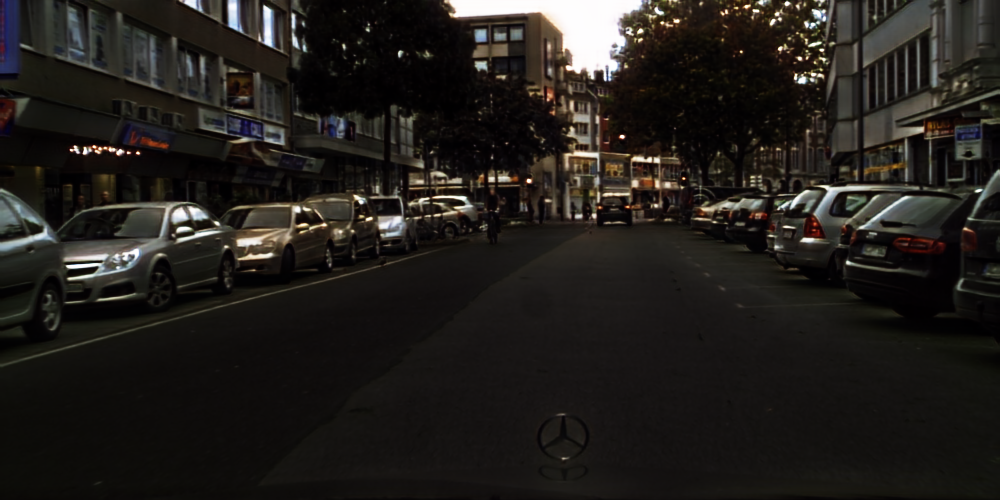}
	\includegraphics[width=0.24\textwidth]{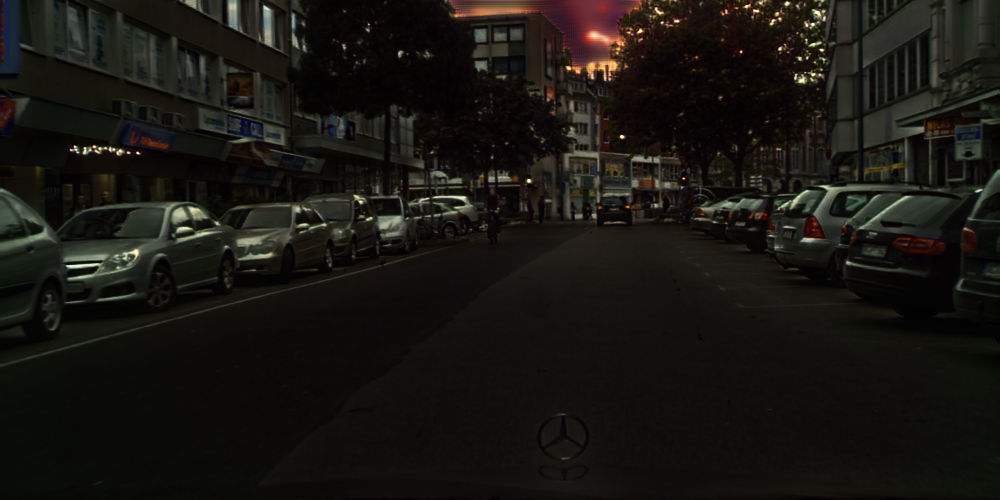} \\
	\includegraphics[width=0.24\textwidth]{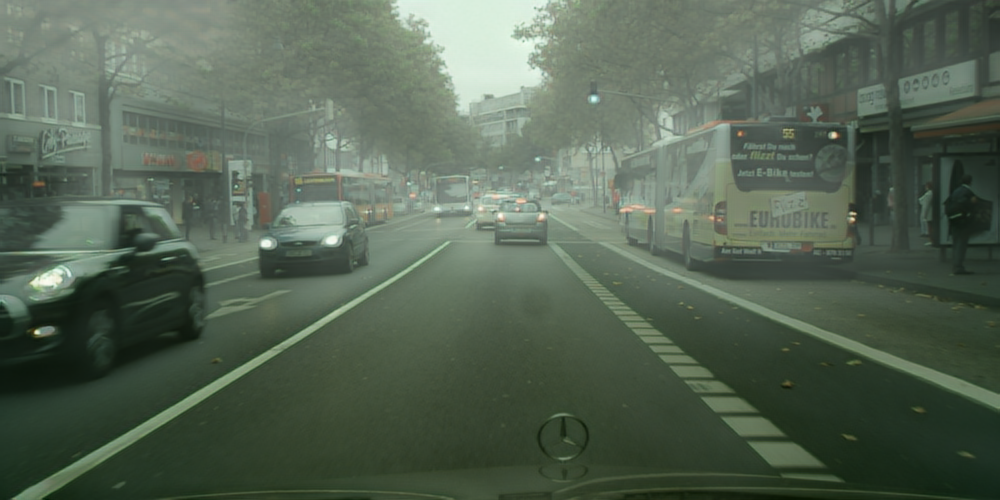}
	\includegraphics[width=0.24\textwidth]{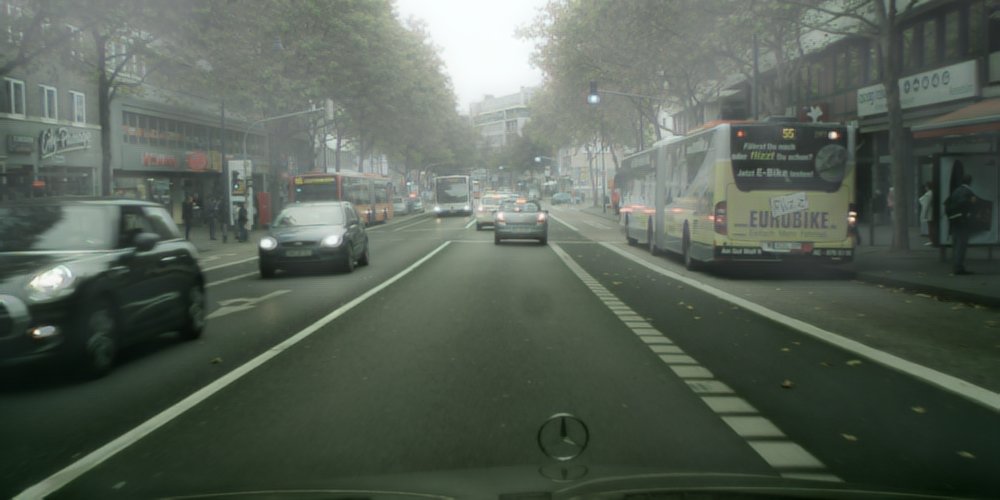}
	\includegraphics[width=0.02\textwidth]{img/white_bar.jpg}
	\includegraphics[width=0.24\textwidth]{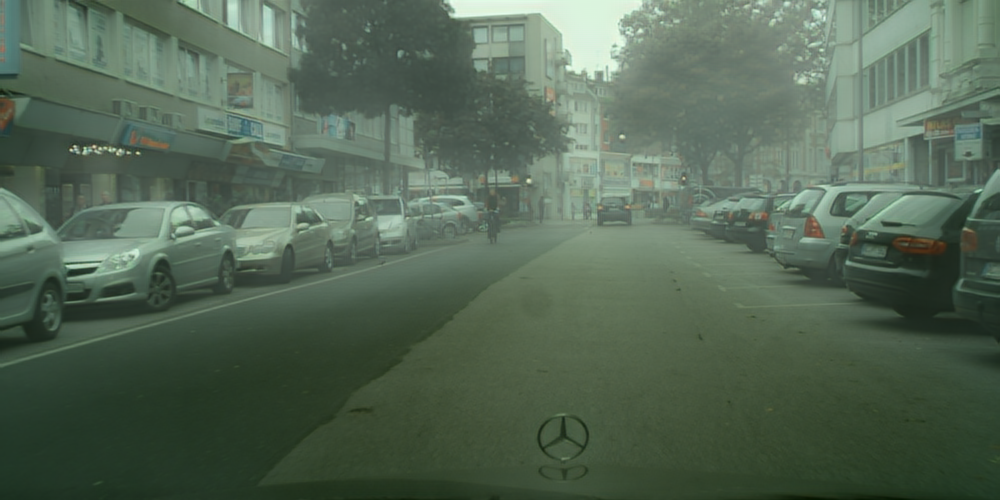}
	\includegraphics[width=0.24\textwidth]{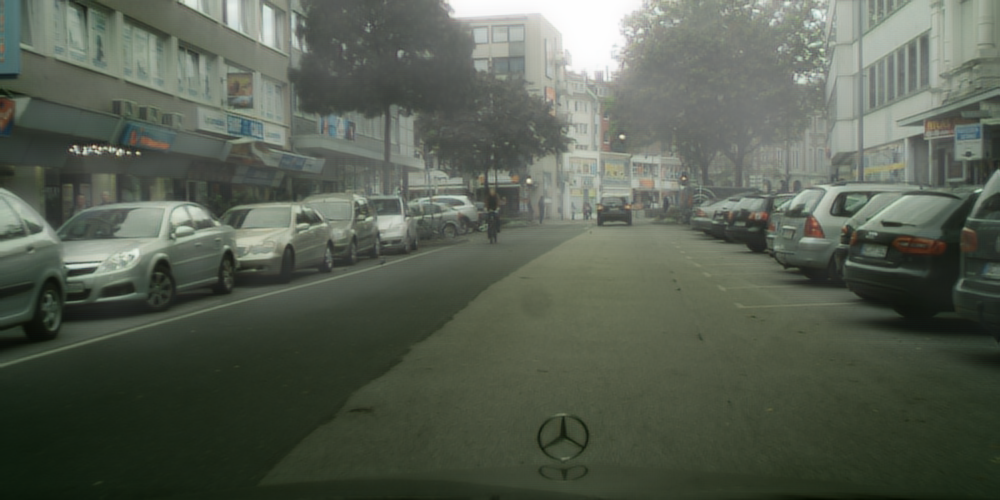} \\
	\caption{Sample target domain images generated by MUNIT models trained using ground truth domain labels and using our proposed unsupervised domain clustering method.}
	\label{fig:proposed_gen_img}
\end{figure*}

\begin{table*}[!htpb]
	\begin{center}
		\begin{tabular}{ c | c c c c c c c c | c}
			AP (\%)  & bicycle  & bus   & car  &  motorcycle &  person & rider & train & truck & mAP \\
			\hline
			baseline   & 37.52  & 52.39  & 61.60  & 33.43  & 44.27  & 46.29  &  27.21  & 31.52  & 41.78  \\ 
			supervised & 36.83  & 52.65  & 62.63  & 30.80  &  43.84  & 45.42  & 38.50  & 32.21  & 42.86 \\
			naive mix  & 34.76  & 43.16  & 60.28  & 33.46  &  43.54  & 43.69  & 27.84  & 26.07  & 39.10  \\  
			gt domain  & 36.88   & 46.17   & 62.35   & 33.03  &  43.27 &  45.95 & 36.83 & 29.64 & 41.76 \\
			proposed   & 36.96  & 48.47  & 62.09  & 32.79  & 42.80  & 45.12  & 36.00  & 26.65  & 41.36
		\end{tabular}
	\end{center}
	\caption{Object detection results on Cityscapes \cite{Cordts2016Cityscapes} dataset.}
	\label{tab:obj_cityscapes}
\end{table*}

\begin{table*}[!htpb]
	\begin{center}
		\begin{tabular}{ c | c c c c c c c c | c}
			AP (\%)  & bicycle  & bus   & car  &  motorcycle &  person & rider & train & truck & mAP \\
			\hline
			baseline   & 12.66  & 35.20  & 47.56  & 14.35  & 26.35  & 25.87  &  8.77 & 21.45  & 24.03  \\ 
			supervised & 31.41  & 48.55  & 61.66  & 28.17  &  40.04  & 40.59  & 25.16  & 23.60  &  37.40 \\
			naive mix  & 21.00  &  42.09  & 52.09  & 15.81  &  31.36  &  28.86  & 28.27  & 20.66  & 30.02 \\  
			gt domain  & 27.38   & 38.16  & 59.01   & 23.34  &  35.14  & 34.63  & 41.09  & 23.99  & 35.34 \\ 
			proposed   & 24.60  &  38.82  & 53.25  &  16.69  & 32.76  & 30.62  & 25.94  & 19.84  & 30.31
		\end{tabular}
	\end{center}
	\caption{Object detection results on rainy Cityscapes \cite{Li_2019_CVPR} dataset.}
	\label{tab:obj_rainy}
\end{table*}

\begin{table*}[!htpb]
	\begin{center}
		\begin{tabular}{ c | c c c c c c c c | c}
			AP (\%)  & bicycle  & bus   & car  &  motorcycle &  person & rider & train & truck & mAP \\
			\hline
			baseline   &  30.66  & 35.38  & 52.36  & 20.75  & 35.38  & 39.27  & 31.15  & 22.63  &  33.45 \\ 
			supervised &  36.33  & 46.60  & 60.81  & 27.74  & 40.91  & 44.13  & 32.64  & 24.16  &  39.16 \\
			naive mix  &  29.80  & 42.87  & 52.98  & 16.38  & 35.96  & 38.72  & 27.88  & 26.58  &  33.90 \\  
			gt domain  &  34.64  & 48.10  & 58.60  & 25.32  & 39.39  & 44.27  & 39.55  & 24.60  &  39.31 \\ 
			proposed   &  33.72  & 48.34  & 58.51  & 25.65  & 38.18  & 41.97  & 16.63  & 27.39  &  36.30
		\end{tabular}
	\end{center}
	\caption{Object detection results on night Cityscapes \cite{night} dataset.}
	\label{tab:obj_night}
\end{table*}

\begin{table*}[!htpb]
	\begin{center}
		\begin{tabular}{ c | c c c c c c c c | c}
			AP (\%)  & bicycle  & bus   & car  &  motorcycle &  person & rider & train & truck & mAP \\
			\hline
			baseline   &  31.88  & 24.07  & 39.33  & 23.42  & 33.05  & 38.47  & 12.04  & 16.25  & 27.31 \\ 
			supervised &  37.15  & 43.29  & 60.00  &  28.05  & 41.97  & 43.32  & 28.82  & 22.58  & 38.15 \\
			naive mix  &  29.54  & 33.50  & 44.25  & 22.72  &  34.68  & 36.70  & 16.94  & 15.92  & 29.28  \\
			gt domain  &  33.36  & 35.22  & 49.77  &  26.32  & 35.43  &  42.81  & 27.45  & 24.58  & 34.37 \\ 
			proposed   &  33.67  & 34.06  & 48.29  &  23.13  & 34.99  & 42.27  & 20.47  & 18.99  & 31.98
		\end{tabular}
	\end{center}
	\caption{Object detection results on foggy Cityscapes \cite{SDHV18} dataset.}
	\label{tab:obj_foggy}
\end{table*}

\section{Experiments}
\label{sec:experiments}

\subsection{Implementation details}
We test our proposed method on tensorflow \cite{tensorflow2015-whitepaper} version implementation \cite{faster-rcnn-github} of Faster R-CNN \cite{Ren2016} with ResNet-101 \cite{He_2016_CVPR} as backbone feature extractor, and use open source implementation of MUNIT \cite{Huang_2018_ECCV} and scikit-learn \cite{scikit-learn} implementation of t-SNE \cite{t-sne}, Silhouette Coefficient \cite{ROUSSEEUW198753} and k-means \cite{1056489,ilprints778}. For all experiments, we report mean average precision (mAP) with a threshold of 0.5 for evaluation.

\subsection{Datasets}
We use the urban scene Cityscapes dataset \cite{Cordts2016Cityscapes} as source domain $\mathcal{S}$, from which three different target domains (weather conditions) are synthesized: foggy \cite{SDHV18}, rainy \cite{Li_2019_CVPR} and night \cite{night}.  Each domain has 2,975 training images, and 500 test images.  There are eight categories of annotated objects for detection \ie bicycle, bus, car, motorcycle, person, rider, train and truck.  Sample images of four domains are shown in Figure~\ref{fig:samples_images}.

\subsection{Target domain image generation and unsupervised domain clustering}
As mentioned in Sec.~\ref{sec:proposed_method}, in the first step we naively treat the mixed target domains as one without distinguishing them, and the resultant $\text{MUNIT}^{\mathcal{T}_{mix}}$ model trained between $\mathcal{S}$ and $\mathcal{T}_{mix}$ cannot generate images that reflects the distribution of $\mathcal{T}_{mix}$ as shown in Figure~\ref{fig:naive_mix_gen_img}.  Though many of the target domain images generated by $\text{MUNIT}^{\mathcal{T}_{mix}}$ do not visually resemble those in $\mathcal{T}_{mix}$ (see Figure~\ref{fig:samples_images}), we found that the AdaIN \cite{Huang_2017_ICCV} parameters extracted from $x^{T} \in \mathcal{T}_{mix}$ by $\text{MUNIT}^{\mathcal{T}_{mix}}$ can be used to separate the distinct target domains.  Figure~\ref{fig:2d_adaIN_params} (better viewed in color) shows 2D t-SNE \cite{t-sne} embedded AdaIN \cite{Huang_2017_ICCV} parameters of $x^{T} \in \mathcal{T}_{mix}$ extracted by $\text{MUNIT}^{\mathcal{T}_{mix}}$.  The samples of rainy, night and foggy images are colored blue, green and red respectively.  It can be seen that the clusters are roughly consistent with different domains, though the confusion between rainy and foggy images exists.  Based on a reasonable guess that there are roughly $2 \sim 4$ different weather conditions and the Silhouette Coefficient \cite{ROUSSEEUW198753} evaluation, the proper number of clusters is found to be 3.  The numerical k-means \cite{1056489,ilprints778} clustering results for $k = 3$ are shown in Table~\ref{tab:clustering}, where each row lists the ingredient of one cluster.  It can be seen that the three different domains are roughly separated, \ie cluster 1 mainly consists of rainy images, cluster 2 of night images and cluster 3 of foggy images.  As described in Sec.~\ref{sec:proposed_method}, once the mixed target domain images are divided into $k$ groups based on their style, we train another $k$ MUNIT models, one $\text{MUNIT}^{\mathcal{T}^{'}_{j}}$ for each source $\mathcal{S}$  and separated target domain $\mathcal{T}^{'}_{j}$ pair, and use these $k$ MUNIT models to transform the annotated source domain images into distinct annotated target domains, obtaining the augmented training dataset.   Figure~\ref{fig:proposed_gen_img} shows sample images generated by $\text{MUNIT}^{\mathcal{T}^{'}_{j}}$ (first, third column) and MUNIT models trained using ground truth domain labels (second, fourth column). For each row, images of the first and second column are synthesized from the same source domain images. This is similar for the third and fourth column. Images of each row correspond to a different target domain.  Compared with Figure~\ref{fig:naive_mix_gen_img}, it can be seen that using proposed method enhances the quality of the generated target domain images, and they are close to that obtained by using ground truth domain labels.

\subsection{Object detection results}
The object detection test results on Cityscapes \cite{Cordts2016Cityscapes}, rainy Cityscapes \cite{Li_2019_CVPR}, night Cityscapes \cite{night} and foggy Cityscapes \cite{SDHV18} are shown in Table~\ref{tab:obj_cityscapes}, Table~\ref{tab:obj_rainy}, Table~\ref{tab:obj_night} and Table~\ref{tab:obj_foggy} respectively.  In each table our `baseline' is the Faster R-CNN \cite{Ren2016} trained on source domain (Cityscapes \cite{Cordts2016Cityscapes}) without any adaptation; the `supervised' model is the Faster R-CNN \cite{Ren2016} supervised trained on all four domains, \ie using object detection annotations of all domains, and its performance can be seen as the upper bound of any unsupervised domain adaptation method; `naive mix' represents the detector trained by Cityscapes \cite{Cordts2016Cityscapes} and $\text{MUNIT}^{\mathcal{T}_{mix}}$ generated target domain images; `gt domain' represents the detector trained on images generated by MUNIT models trained using ground truth domain labels instead of the results of our proposed unsupervised domain classification; the performance of the detector trained by our proposed method is list in the last row.  

Using a style transfer model to augment the training images even in the most naive way proves the robustness of the object detector, so `baseline' is outperformed by any other models.  Comparing `supervised' and `gt domain' it can be seen that if a perfect domain classifier is available, the detection results of a supervised trained detector and a domain adapted detector are almost the same in Cityscapes (Table~\ref{tab:obj_cityscapes}) and night Cityscapes (Table~\ref{tab:obj_night}), and close for the rest two domains.  This shows the effectiveness of our proposed  multi-target domain adaptation framework.  The huge gap between the performance of `naive mix' and `gt domain' shows that if multiple target domains exist, naively treating them as one and applying the domain adaptation method designed for single target domain is far from satisfactory.  Comparing `naive mix' and `proposed' it can be seen that adopting our proposed unsupervised domain classification method improve the detection performance in all domains, and this shows the effectiveness of our proposed method.

\section{Conclusion}
\label{sec:conclusion}
In this paper, without using domain labels, we propose a multi-target domain adaptation method based on unsupervised domain classification.  The object detector trained by our proposed method achieves robust detection performance under different weather conditions.  We also propose to use a style transfer model to extract style features for unsupervised domain classification, this novel method is applicable in a wide range of computer vision tasks.

{\small
\bibliographystyle{ieee_fullname}

}

\end{document}